# A Collaborative Approach using Ridge-Valley Minutiae for More Accurate Contactless Fingerprint Identification

*Ritesh Vyas*, *Ajay Kumar*

Department of Computing, The Hong Kong Polytechnic University
Email: csrvyas@comp.polyu.edu.hk, csajaykr@comp.polyu.edu.hk

**Abstract:** Contactless fingerprint identification has emerged as an reliable and user friendly alternative for the personal identification in a range of e-business and law-enforcement applications. It is however quite known from the literature that the contactless fingerprint images deliver remarkably low matching accuracies as compared with those obtained from the contact-based fingerprint sensors. This paper develops a new approach to significantly improve contactless fingerprint matching capabilities available today. We systematically analyze the extent of complimentary ridge-valley information and introduce new approaches to achieve significantly higher matching accuracy over state-of-art fingerprint matchers commonly employed today. We also investigate least explored options for the fingerprint color-space conversions, which can play a key-role for more accurate contactless fingerprint matching. This paper presents experimental results from different publicly available contactless fingerprint databases using NBIS, MCC and COTS matchers. These consistently outperforming results also help to validate the effectiveness of the proposed approach for more accurate contactless fingerprint identification.

## 1. Introduction

Biometric patterns offer the most reliable signatures to conveniently and securely establish human identities. These signatures are derived from the physiological and/or behavioral characteristics of respective individuals. Among several physiological traits accessible from the human body, finger ridge patterns have been widely employed in law enforcement departments to establish unique personal identity of suspects. The fingerprint features are formed before the birth, have high permanence and use of today's high-computing machines have made their use quite convenient and fast. The pervasiveness of fingerprint authentication can be conjectured by its use in the most ubiquitous devices *i.e.* smartphones. Moreover, there are a wide range of e-business applications like financial transactions or access to secured offices, which have increasingly relied on fingerprint authentication.

However, till date, majority of the fingerprint-based systems use contact-based sensors, to acquire the fingerprint of any subject, *i.e.* these sensors require the subject to make a contact of his or her finger with the platen or surface of the sensors. Such contact-based acquisition poses new challenges relating to user convenience, hygiene and security threats. Contact-based requirement can be a serious threat to hygiene as there is a wide range of diseases, *e.g.* severe acute respiratory syndrome (SARS), which are known to transmit or spread from unintended contacts. Therefore, the user's hygiene become vulnerable in such contact-based acquisition. The leftover or latent impressions on the surface of contact-based sensors not only interfere with the new acquisitions but are also known to pose security threat as these can be lifted to reconstruct spoof fingerprints for the presentation attacks.

Contactless acquisition of fingerprint can address above-mentioned challenges and has increasingly attract-ed attention of researchers and developers [1]. Several references [4], [19] have investigated contactless fingerprint acquisition using mobile phones and specialized setups while many commercial contactless fingerprint sensors have also been recently emerged for deployments. Availability of contactless fingerprint databases [13], [16] have also encouraged much needed further research in this area. Involuntary finger motion during the contactless fingerprint imaging can significantly degrade the matching accuracies and reference [7] has attempted to address such problem. Earlier work [2], [9], [16], [19] in contactless 2D finger-print identification have largely incorporated image segmentation, enhancement or minutiae matching algorithms that have shown promising results for the contact-based fingerprints. Such *direct* use of contact-based fingerprint methods ignores the nature of image formation from con-tactless sensing and is therefore not adequate to utilize *full potential* from contactless fingerprint images. The duality of relationship between minutiae extracted from the contactless fingerprint images and contact-based fingerprints has been quite known [17] and considered during cross-sensor fingerprint matching. However, this paper for the *first time* analyzes such influence in the real contactless fingerprint databases and introduces alternative strategies to achieve significant performance improvements using the popular contact-based fingerprint matchers.

## 2. Contactless Fingerprint Image Formation

Human fingers are known to be a curved 3D structure [5], [18]. However when these 3D surfaces are sensed using a contact-based fingerprint sensors, only its 2D projections are recorded. More precisely, when the subject touches the sensor platen with his or her finger, the topographic high points, which are known as the ridges, are imaged while the low points, known as the furrows or valleys, are not imaged as these are considered as part of the background. Figure 1(a) illustrates the frustrated total internal reflection (FTIR) principle, which is popularly employed in the legacy contact-based fingerprint sensors. It can be observed from this figure that the pixels corresponding to the light rays reflected from the low topographic regions, or the valleys of finger skin surface, have higher intensity (bright) in a fingerprint image. The spatial locations from high areas or ridges are rendered as darker regions in the fingerprint image. The *same* ridge and valley topologies have been *correspondingly* identified in Fig. 1(b).

However the contrast between such gray-levels between the ridge and valley is remarkably different in the contactless acquisition (see Fig. 1(c)) of the *same* finger surface. This difference can be attributed to the varying illumination on different topographic regions of the finger skin where the ridges are rendered as darker or brighter areas in those regions. Contactless acquisition is commonly followed by grayscale representation, or the binarization for the finger images, so that their appearance is similar to their legacy contact-based counterparts. Such second-order representation for the contactless fingerprints can significantly degrade the system's matching accuracy. A gray-level representation can be attributed to the variation in the illuminating angle, because of which the ridges are sometimes rendered as brighter and sometimes as darker than the adjacent valleys. One such instance is illustrated in Fig. 1(d). In summary, the brighter and darker portions of a ridge in a contactless image are not consistent with ridge and valley. Instead, they are consistent with the opposite flanks of ridges. Such *polarity reversal* effect is quite common and can be attributed to the inter-action of incident illumination, with respect to the 3D ridge-valley structure [8], [11], during the contactless imaging.

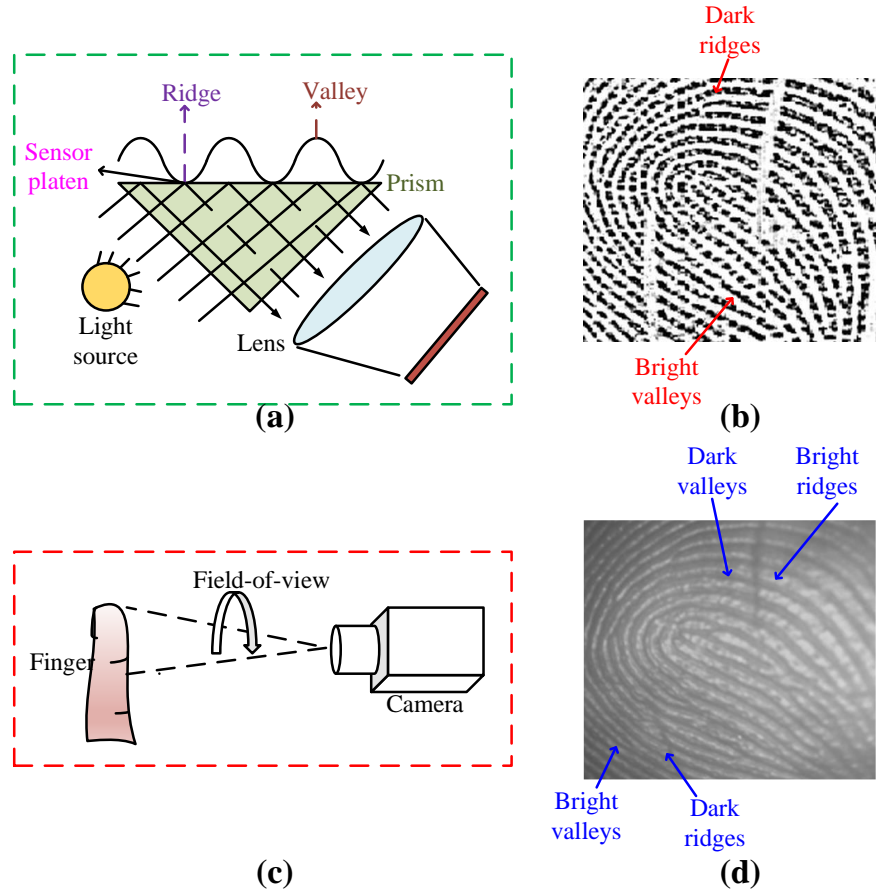

**Figure 1**: Illustration of legacy and contactless fingerprints. (a) FTIR in legacy fingerprints, (b) Legacy (contact-based) fingerprint, with regular dark ridges and bright valleys, (c) Contactless acquisition, (d) Contactless fingerprint (with ridges and valleys in varying polarities).

These gray-level alterations can be easily observed from the *raw* fingerprint images acquired in contactless manner. However, if the images are represented as grayscale, or even binarized to detect the potential minutiae points, this polarity reversal effect is completely lost. This is also the key reason for poor performance when the contactless images are matched against the legacy contact-based fingerprints. Our detailed observations reveal that the minutiae expected to be matched from contactless and contact-based images do not appear at same or similar positions. The ridge endings of contact-based fingerprint become ridge bifurcations in the contactless fingerprint and vice-versa. An instance of such changes in the type of minutiae is illustrated from a real finger image sample in Fig. 2 below.

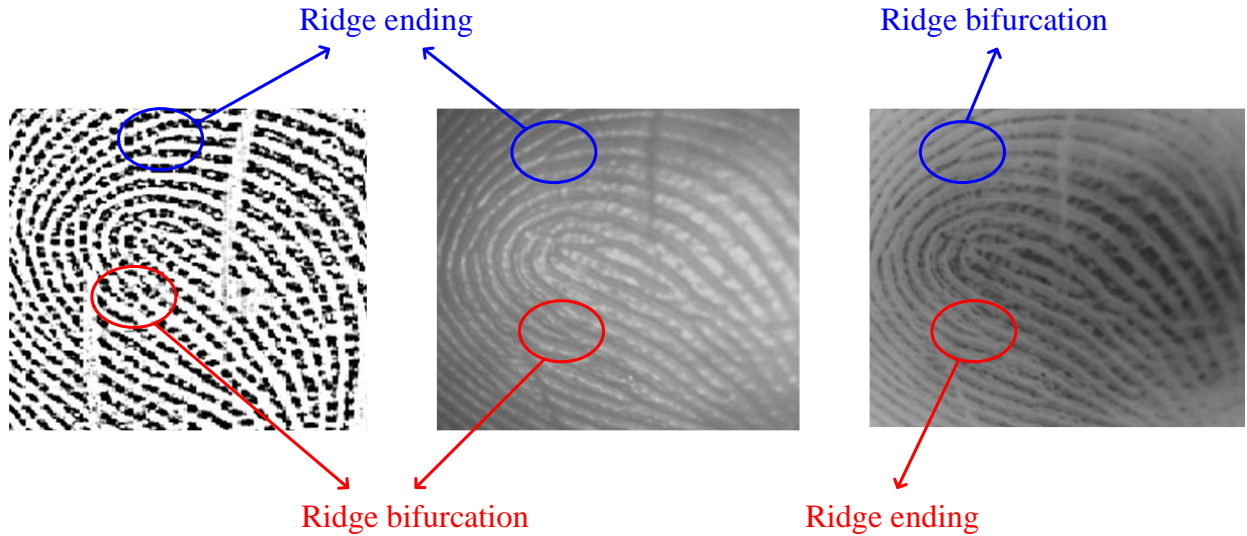

**Figure 2**: Differences in the localization of fingerprint minutiae from (left) con-tact-based or legacy fingerprint image, (centre) contactless finger-print image, and respective (right) inverted fingerprint image

Such image formation mechanism during the *contactless* fingerprint acquisition motivated us to investigate additional information for the localization of minutiae from the input fingerprint image. Unlike the legacy contact-based fingerprint, in which the valley information is lost as it is considered as the part of background information, the contactless fingerprint imaging *simultaneously* recovers the ridge and valley inormation. However, this joint information is embedded with low-contrast as 3D ridges are imaged under uneven illuminations and under reflections from multiple ridges. In order to reveal additional minutiae from the otherwise dark portions of the contactless image and exploit the polarity alternation notion, we also investigate image transformations to precisely recover the minutiae. This work presents such systematic investigation, using publicly available contactless fingerprint image databases, and introduces possible solutions to address significant degradation in performance while matching contactless fingerprint images.

## 3. Accurate Minutiae Detection from Contactless Fingerprints

The difference between the raw and its inverted contact-less fingerprint images can be explicitly observed from the sample images in Fig. 3. In this figure, the top and bottom rows show the detected minutiae from raw and inverted contactless fingerprints, respectively. As can be observed from these images that most of minutiae detected from the raw contactless fingerprint are also present in the corresponding inverted fingerprint, just with a change in minutiae type. This explains for bifurcations of raw fingerprint appearing as terminations in the inverted images and vice-versa [20]. More importantly, it can also be clinched upon careful inspection of these images that many spurious minutiae, which are falsely detected from the raw fingerprints, remain *undetected* in their inverted counterparts. Additionally, the inverted fingerprint *facilitates* the detection of *additional* cores and deltas, which otherwise remain undetected from the raw fingerprints.

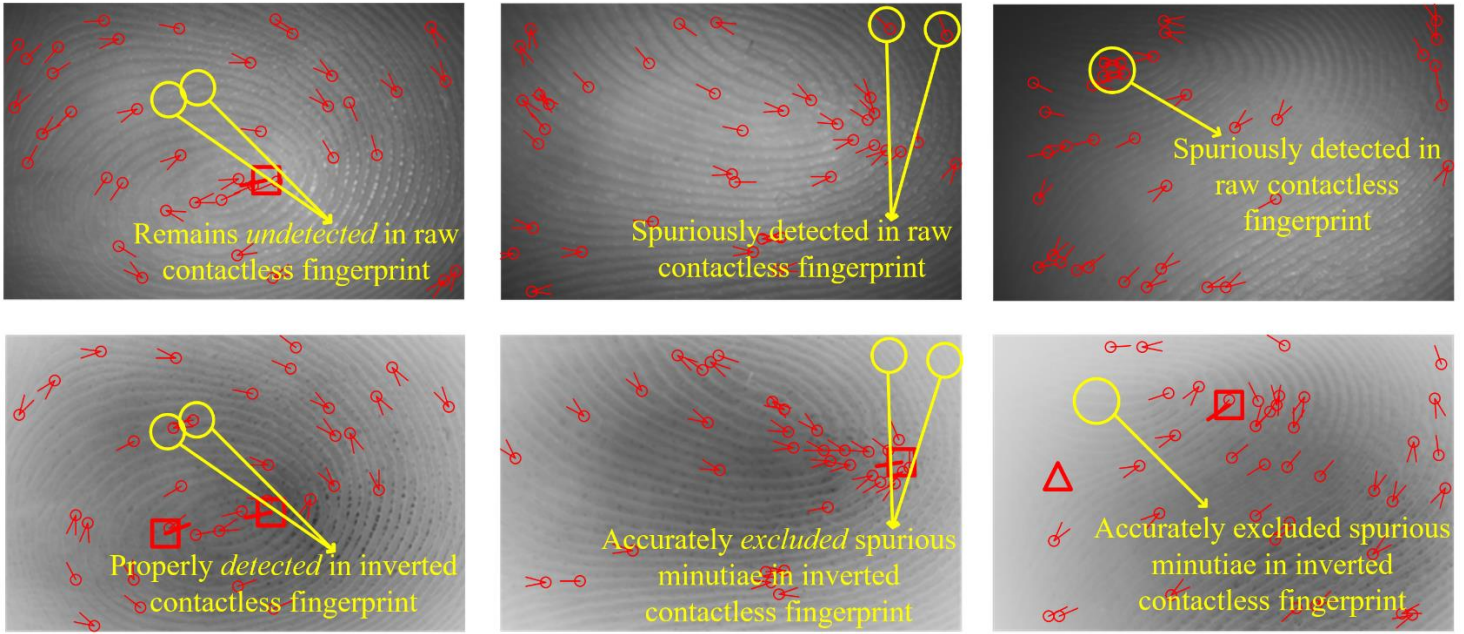

**Figure 3**: Minutiae detection in (top row) raw contactless fingerprints and (bottom row) inverted contactless fingerprints.

For instance, the first image sample in the top row of Fig. 3 shows a sample fingerprint having two cores (i.e. u-shaped ridges) in the lower half of the image. However, the popular commercial off-the-shelf (COTS) tool, namely *VeriFinger* [12], is only able to detect single core (shown by red-colored rectangle), while the other one remains unnoticed because of poor focus of the image. However, when the inverted fingerprint is provided as input to the same COTS tool, both the cores get detected (highlighted by two red-colored rectangles in the first image in bottom row). Similar observations can be made from the second and third image of Fig. 3, where the cores of fingerprints are detected only when their inverted version is presented as input (see the red-colored rectangles in second and third images of bottom row). Moreover, the last fingerprint in top row Fig. 3 possesses a delta (*i.e.* a Y-shaped ridge), which can be observed with the bare eyes on the raw contactless image. However, this delta remains undetected because of the poor focus on this part of the finger, which can be attributed to the curved 3D profile of human fingers. On the other hand, the COTS tool was able to detect this delta from the inverted fingerprint image (red-colored triangle in last fingerprint of bottom row).

## 3.1 Grayscale Representation for Colored Contactless Fingerprints

In order to study the effect of different grayscale representation, the unprocessed fingerprint images from PolyU Contactless to Contact-based fingerprint database [13] were selected. These unprocessed fingerprints are available in RGB color representation and have larger size (i.e. 1400 × 900 pixels) than processed ones. The PolyU database [13] also provides a processed contactless fingerprint sub-dataset comprising of grayscale 350 × 225 pixel images. These ordinary grayscale images are generally formed by weighted combination of linear intensities observed in individual channels. Mathematically, such conversion of RGB images to grayscale images can be directly achieved as follows:

$$I_{Ordinary} = 0.3 I_R + 0.59 I_G + 0.11 I_B \quad (1)$$

where, $I_R$, $I_G$ and $I_B$ are the linear intensities in the red, green and blue channels, respectively, in the colored images. However, there are other grayscale representations, which can reveal additional details from the output images. In order to provide more illustrious case of study, we adopted the Luma grayscale representation [14] as an alternative to the grayscale representation in eq. (1). The Luma grayscale representation is believed to be more perceptually accurate grayscale representation, as it employs non-linear gamma corrected versions [15] of all channels in colored images, rather than their linear intensities. Mathematically, it can be expressed as follows:

$$I_{Luma} = 0.2126 \, I'_R + 0.7152 \, I'_G + 0.0722 \, I'_B \quad (2)$$

where, $I'_R$, $I'_G$ and $I'_B$ are the gamma-corrected versions of red, green and blue channels, respectively. The Luma representation of colored images can enable larger contrast as compared to the ordinary grayscale images, which facilitates the enhanced detection of minutiae features. This observation can also be noted from the instances of grayscale images shown in Fig. 4. It is not difficult to observe that Luma grayscale images have increased contrast as compared to the ordinary grayscale images. It is evident from the first and second sample in Fig. 4, that the delta singularity is appropriately detected in Luma grayscale images, while it is not detected in the ordinary grayscale images. Moreover, a careful visual inspection reveals the presence of many spurious minutiae, which are erroneously detected in the ordinary grayscale representation, gets suppressed in the Luma grayscale represented (refer to last column images of Fig. 4) images.

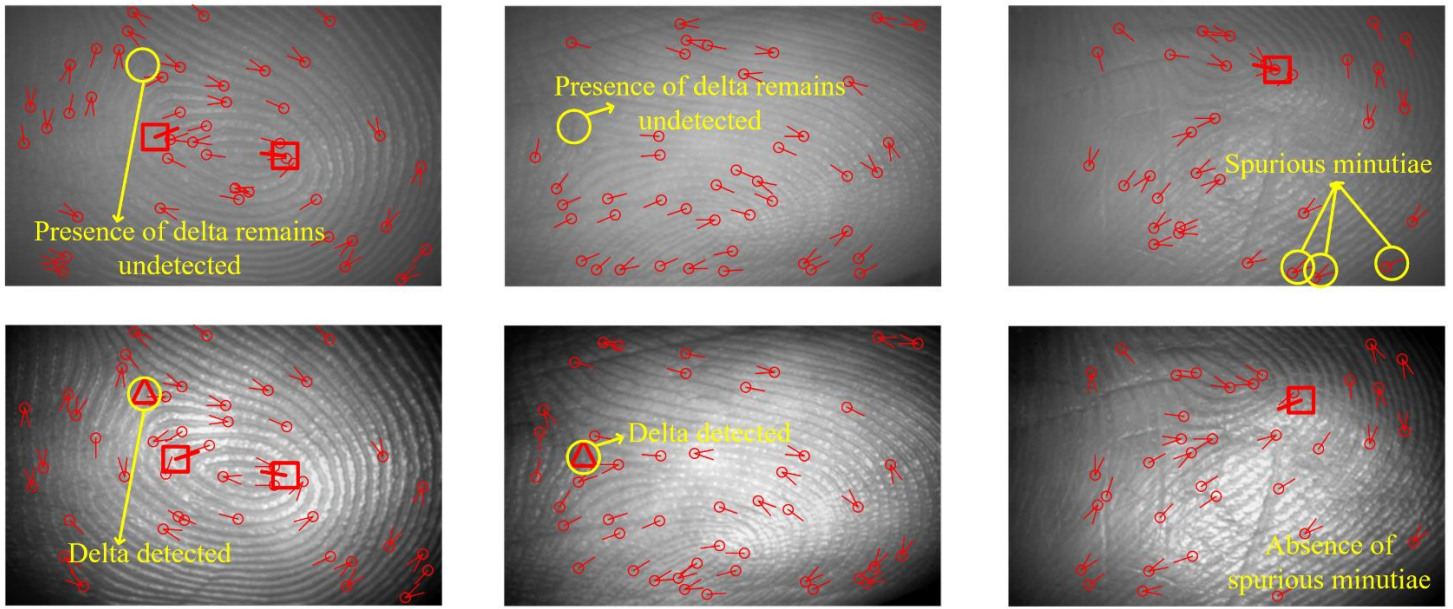

**Figure 4**: Different grayscale representations: Ordinary grayscale in top-row images and Luma grayscale in bottom-row images.

Therefore, in order to add a new dimension to the enhancement for contactless fingerprint recognition, we performed experiments with Luma grayscale contactless fingerprint images, which were obtained by converting the unprocessed (colored) images of PolyU database to Luma grayscale and reducing their dimension to be same as that of the processed contactless fingerprints in the database i.e $350 \times 225$ pixels. Thereafter, the recognition experiments were also performed on the inverted versions of Luma grayscale images to present a comprehensive evaluation in the currently selected framework. These experimental results, with both the Luma grayscale images and their inverted counterparts, are discussed on the next section.

## 4. Experiments and Results

*Databases*

The present work comprises of experiments on two popular contactless fingerprint databases. The first database is the PolyU Contactless to Contact-based fingerprint database [13], which provides 2016 contactless 2D fingerprints and their corresponding 2D contact-based 2016 fingerprints. These fingerprints were captured from 336 clients from the staff and students at the university. Each client has six fingerprint images, both in the contactless and contact-based acquisition setup. In our experiments, we use all the 2016 contactless fingerprint images. Another contactless fingerprint database used in this work is the Benchmark 2D/3D Fingerprint Database publicly available from [16], which provides contactless and contact-based fingerprints from 1500 different fingers. The database comprises of at least two contactless fingerprint samples and four contact-based fingerprint samples. The acquisition of this database was completed at three different universities in Australia. We employed images from 1000 fingers of this database in our experiments. Each of these fingers have two contactless fingerprints, which resulted in a total of 2000 images. Sample images from both of these public databases are shown in Fig. 5.

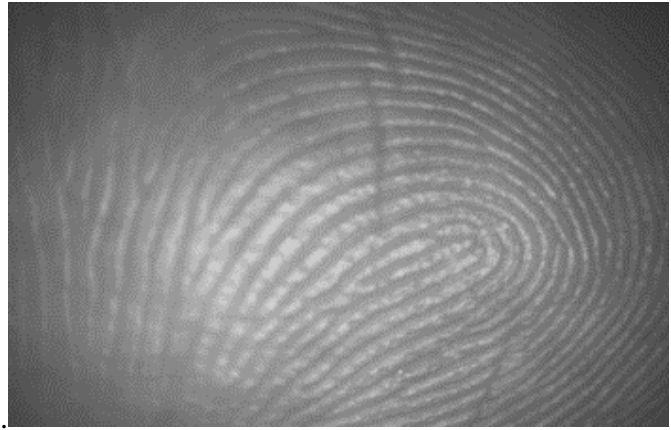    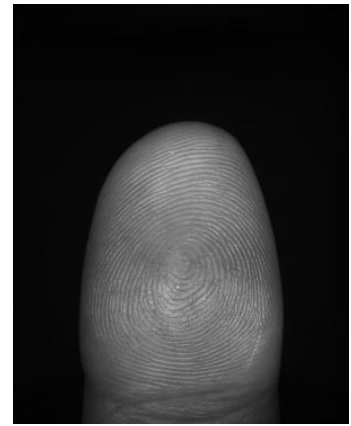

(a)                                                                                                         (b)

**Figure 5**: Samples from both the contactless fingerprint images database (a) PolyU Contactless to Contact-based database [13], (b) benchmark 2D/3D contactless fingerprint database [16].

*Evaluation protocol and performance metrics*

All experiments were performed using all-to-all matching protocol, in which every single fingerprint is matched against all other fingerprints in the database. This protocol, being the *most* challenging biometric evaluation protocol, yields a large number of scores for both the employed databases. The matchings for PolyU Contactless to Contact-based fingerprint database generate 5,040 genuine and 20,26,080 imposter scores. On the other hand, numbers of genuine and imposter scores for the 2D/3D benchmark contactless database are 1,000 and 19,98,000, respectively.

Three state-of-the-art fingerprint matchers are employed for the performance evaluation. First is a popular fingerprint matcher NBIS (NIST Biometric Image Software) [17]. The second matcher is minutiae cylinder code (MCC) [18] while third matcher is the commercial off-the-shelf (COTS) matcher, namely *VeriFinger* from Neurotechnology [12]. Performances of the raw contactless fingerprints and their inverted versions are comparatively evaluated using the common performance metrics like receiver operator characteristics (ROC), equal error rate (EER), false acceptance rate (FAR), genuine acceptance rate (GAR). Performances from all three matchers are detailed in following.

## 4.1 NBIS Matcher

The NBIS [17] is an open-source fingerprint matcher provided by NIST. This matcher uses two key modules, namely MINDTCT and BOZORTH3, for the minutiae detection and template matching respectively. It is open source, complies with the several standards, and employed widely in a range of law-enforcement applications. Therefore, it is employed to validate the effectiveness of our approach for the contactless fingerprint recognition. This matcher generates similarity scores while matching two minutiae templates. The maximum score obtained after matching intra-class fingerprint template lies in the interval of 300-325.

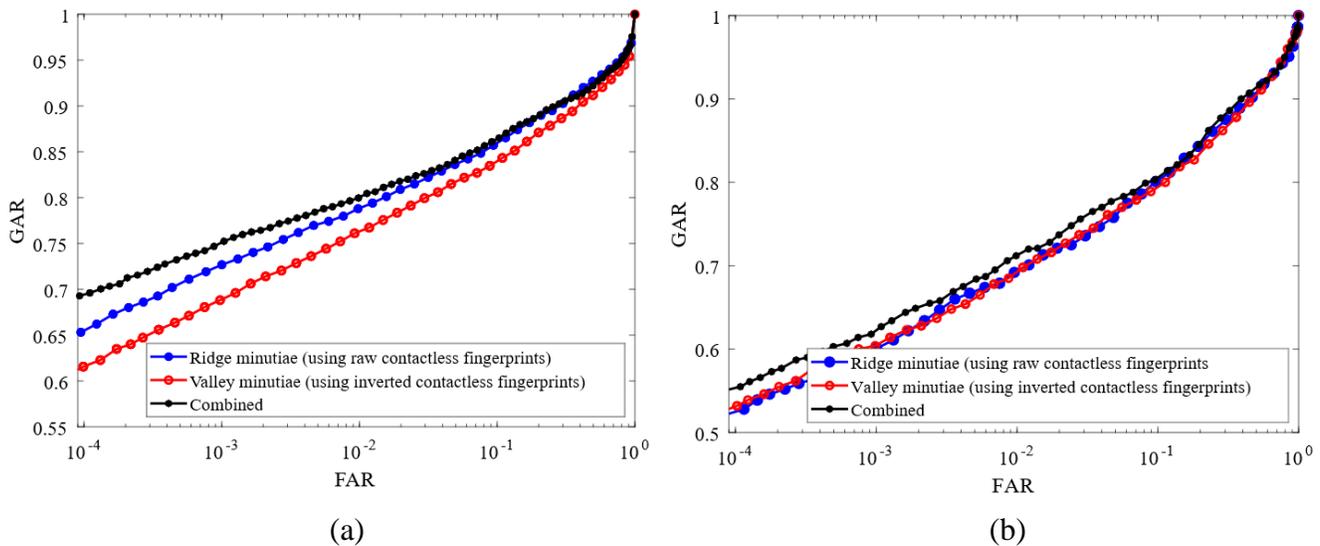

(a)          (b)

**Figure 6**: ROC curves for ridge and valley minutiae matching and combined authentication with NBIS on (a) PolyU contactless fingerprint database and (b) benchmark contactless fingerprint database.

Our experimental results on PolyU contactless fingerprint dataset, using NBIS matcher, are illustrated in Fig. 6(a). The corresponding ROC curves illustrate that NBIS matcher achieves an EER of 13.33% and GAR (@FAR=0.01%) of 65.52% for legacy contactless fingerprints (i.e. with conventional ridge minutiae) in PolyU database. On the other hand, with the usage of inverted contactless fingerprints (i.e. valley minutiae), the observed EER is 14.14% and GAR (@FAR=0.01%) is 61.61%. Hence, it can be inferred that the NBIS matcher can achieve better results with the ridge-minutiae template matching rather than with the corresponding valley-minutiae based template matching. However, it is equally important to note that combination of scores from both scenarios of matching reflects the complementary nature of the inverted fingerprints. The experiments with NBIS on PolyU database also reveal that the matching accuracy from the normal contactless fingerprints is at par with that of the inverted fingerprints. The performance improvement from the combination of two template representations is significant and validates our arguments in section 3.

The performance of NBIS matcher on the contactless fingerprint database from [16] is illustrated using the ROC curves in Fig. 6 (b). The values of performance metrics, the EERs of raw and inverted contactless fingerprints are found to be 16.25% and 17.71%, respectively. On the other hand, the GAR (@FAR=0.01%) for both the ridge and valley minutiae matching scenarios are 52.49% and 53.13%, respectively. From these values, it can be argued that matching the raw fingerprint images offers better EER, but worse GAR, when compared to those from matching the inverted fingerprint images. However, collaboration or the combination of two scores results in improved performance and can also be observed from the corresponding ROC curve.

## 4.2 MCC Matcher

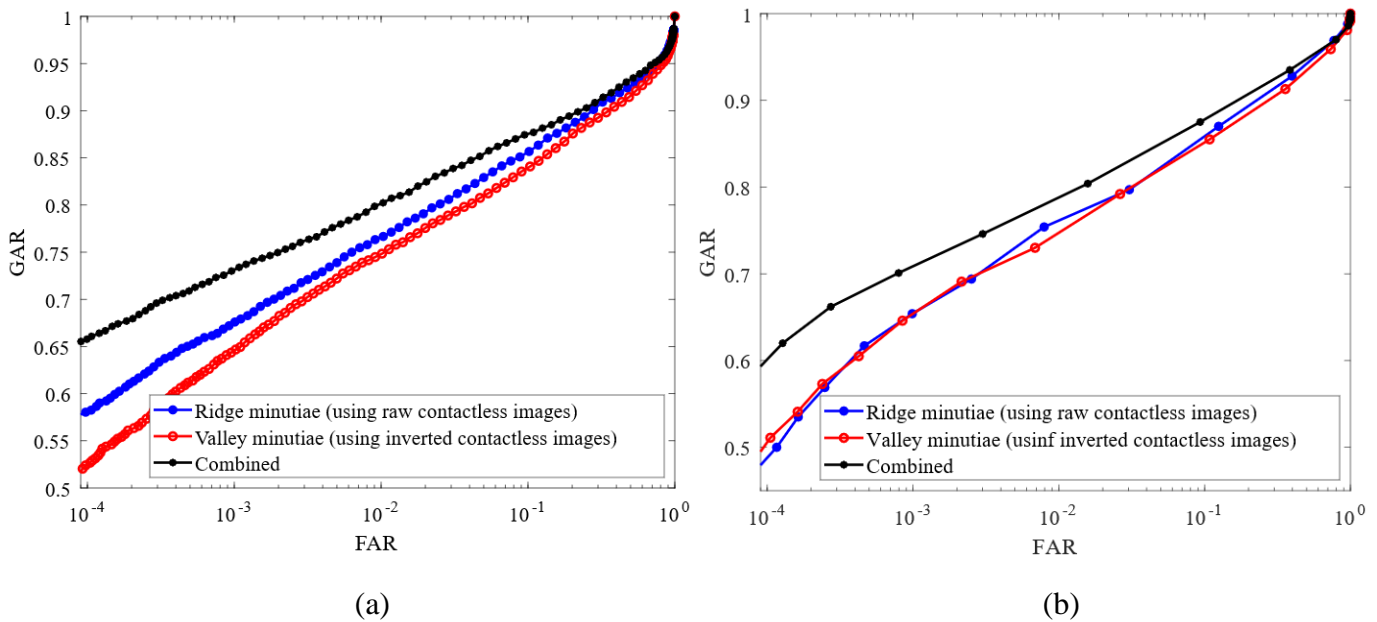

(a)           (b)

**Figure 7**: ROC curves for ridge and valley minutiae matching and their combination, using MCC, for (a) PolyU contactless fingerprint database and (b) benchmark contactless fingerprint database.

The Minutiae Cylinder Code (MCC) [18] is widely considered as a state-of-art fingerprint matcher and therefore also utilized in our work. It is also an open-source fingerprint matcher which firstly computes the local similarities between minutiae based on the 3D structure (cylinder) using the minutiae distances and angles. Eventually, these local similarities are consolidated to form a global score.

The performance of the MCC matcher for PolyU Contactless database using the ROC curves is shown in Fig. 7 (a). From these ROC curves, it is apparent that the MCC matcher generates inferior performance for the inverted contactless fingerprints as compared to the raw contactless images. The EERs obtained using the raw and inverted fingerprint frameworks are 13.25% and 14.07%, respectively. While, the GAR (@FAR=0.01%) for both the above cases are obtained as 58.12% and 52.52%, respectively.

The ROC curves for the contactless benchmark database [16] using the MCC matcher are shown in Fig. 7 (b), which indicates that the inverted contactless fingerprints offer slightly better performance than the raw contactless fingerprints. Considering the performance metrics, the EER values for raw and inverted fingerprints are 12.74% and 12.66%, respectively. The GARs (@FAR=0.01%) for both these cases were 48.82% and 50.65%, respectively. It can however be observed that a notable performance improvement can be achieved, both in EER and GAR, when the individual scores (refer to section 4.4 for more details) are simultaneously employed.

## 4.3 COTS Matcher

A commercial off-the-shelf (COTS) fingerprint matcher, namely *VeriFinger* from *Neurotechnology* [12], was also employed to corroborate the usefulness of inverted fingerprints in improving the contactless fingerprint recognition capabilities. This commercial matcher tool is known to perform excellently in minutiae extraction and matching. The matching of minutiae templates through using this matcher generates similarity scores, which can have values up to infinity. However, the maximum similarity score, generated for matching a fingerprint with itself was observed to be roughly in the interval of 2000-2700 for the PolyU Contactless to Contact-based fingerprint database and 2900-3500 for the contactless 2D/3D benchmark fingerprint database [16].

The ROC curves for the original (or ridge minutiae matching) and inverted (or valley minutiae matching) contactless fingerprint images of the PolyU database are illustrated in Fig. 8 (a). The ROC plots clearly validate the improvements in the contactless fingerprint matching performance with the inverted contactless fingerprints. The verification experiments with inverted versions of contactless fingerprints have clearly outperformed those with the original contactless fingerprints. In terms of performance metrics, experiments with the original contactless fingerprints yield an EER of 0.81% and GAR (@FAR=0.0001%) of 91.11%. On the other hand, experiments with inverted contactless images generated EER of 0.29% and GAR (@FAR=0.0001%) of 98.68%. These values exhibit an improvement of 64.19% and 8.30%, respectively, when compared to their counterparts in the experiments with original images. This significant improvement of performance metrics clearly validates the consideration of polarity alterations in contactless acquisition of fingerprint images.

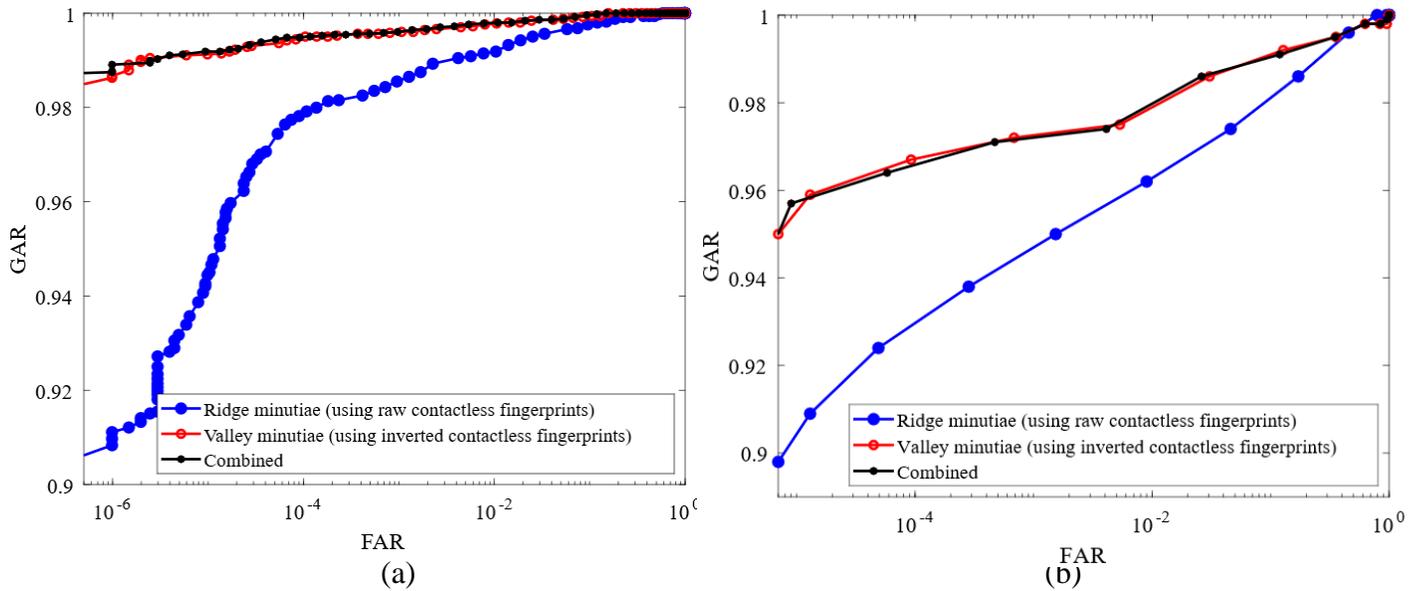

**Figure 8**: ROC curves for ridge and valley minutiae matching and combined authentication using COTS: (a) PolyU contactless fingerprint database, (b) benchmark contactless fingerprint database.

The same set of verification experiments was also performed using benchmark database [16] as for earlier cases. The performance metrics for these experiments (i.e. EER and GAR@FAR=0.001%) for original contactless fingerprints are 3.60% and 90.43%, respectively. On the other hand, observed EER and GAR (@FAR=0.001%) values for the inverted fingerprint images from the same database are 2.23% and 95.52%, respectively. Therefore, these experiments with inverted images show 38.05% and 5.62% improvement in the performance metrics, as compared with those from the experiments using the raw contactless fingerprint images. The corresponding ROC curves are illustrated in Fig. 8 (b). The experimental results on a popular COTS matcher, with known implementation details, also support the merit in the joint use of ridge and valley minutiae matching. The plausible reason for relatively higher EER's from this database [16] can be attributed to the *nature* of fingerprint images in this database. The fingerprints in this database have higher pose variations and with higher distance between the finger and sensor as can also be observed from image sample in Fig. 5 (right) in comparison to the fingerprint sample in PolyU database (Fig. 5 (left)).

It is reasonable to argue that when the matching performance from the inverted contactless fingerprint images is comparable or even *inferior* to those from their legacy counterparts, it can still provide complementary details to enhance the performance from the legacy contactless fingerprint images. This argument is also well supported by the ROC curves illustrated in Figs. 6-8 on *public* contactless fingerprint databases. The key focus of this paper is on investigating the polarity reversal or alterations observed in the common contactless fingerprint images or databases for more accurate performance. A careful observation of the ROC curves of Figs. 6-8, indicated that the fusion of scores from the raw and inverted contactless fingerprints images almost always results in noticeable performance improvements. Such improvements in performance metrics as a result of simultaneous use of these two representations is summarized in Table 1. The results summarized in this table clearly indicates that the minutiae information furnished from the inverted contactless fingerprints can surely aid to improve performance from the raw contactless fingerprints.

## 4.4 Gray Level Transformation

The second part of experimentations were performed to investigate the least-explored notion of effective color-to-grayscale transformation which can play a vital role in localizing key-minutiae and enhancing performance for matching the contactless fingerprint images. One of main advantages of contactless fingerprints is the enhanced user convenience associated with their acquisition with mobile phones which generates colored images. Therefore, effective grayscale representation is expected to enable accurate detection of legitimate minutiae and the suppression of spurious minutiae during template extraction.

By observing the improved performance from Luma grayscale fingerprints, largely due to the appropriate facilitation in the detection of minutiae as shown from samples in Fig. 4, it can be concluded that this grayscale space can lead to the improved contactless fingerprint recognition performance. We therefore performed additional experiments to ascertain this possibility. The ROC curves for these contactless fingerprint verification experiments, using the ordinary and Luma grayscale images, are shown in Fig. 9. In this figure, the ROC curves corresponding to the ordinary and Luma grayscale images are illustrated using the solid and dashed lines, respectively. It is apparent from Fig. 9 that the performance using the Luma grayscale images are noticeably superior over the ordinary grayscale representations.

**Table 1**: Summary of iimprovements in performance metrics from simultaneous ridge and valley minutiae representations.

| Databases \ Matchers | | NBIS | | | | | MCC | | | | | COTS | | | | |
|---|---|---|---|---|---|---|---|---|---|---|---|---|---|---|---|---|
| | | Ridge | Valley | Combined | Comparison w.r.t. | | Ridge | Valley | Combined | Comparison w.r.t. | | Ridge | Valley | Combined | Comparison w.r.t. | |
| | | | | | Ridge | Valley | | | | Ridge | Valley | | | | Ridge | Valley |
| PolyU Contactless to Contact-based Database | EER (%) | 13.33 | 14.14 | 12.50 | ↑6.22% | ↑11.59% | 13.25 | 14.07 | 11.96 | ↑9.73% | ↑14.99% | 0.81 | 0.29 | 0.26 | ↑67.90% | ↑10.34% |
| | GAR (%) | 65.52 | 61.61 | 69.43 | ↑5.96% | ↑12.69% | 58.12 | 52.52 | 65.82 | ↑13.25% | ↑25.32% | 91.11 | 98.68 | 98.91 | ↑8.56% | ↑0.23% |
| Multiview Contactless Fingerprint Database | EER (%) | 16.25 | 17.71 | 16.00 | ↑1.54% | ↑9.66% | 12.74 | 12.66 | 10.91 | ↑14.36% | ↑13.82% | 3.60 | 2.23 | 1.45 | ↑59.72% | ↑34.97% |
| | GAR (%) | 52.49 | 53.13 | 55.35 | ↑5.44% | ↑4.18% | 48.82 | 50.65 | 60.21 | ↑23.33% | ↑18.87% | 90.43 | 95.52 | 95.74 | ↑5.87% | ↑0.23% |

The verification experiments performed with the NBIS matcher show a notable improvement with Luma grayscale images as compared to the ordinary grayscale images. These ROC curves using the NBIS matcher are shown in Fig. 9 (a). The EERs achieved using the original and inverted Luma grayscale images are 6.70% and 6.81%, respectively. These values are improved by 49.73% and 51.84% when compared with the EERs achieved from NBIS matcher for the ordinary grayscale fingerprint images. The GARs (@FAR=0.01%) for raw and inverted Luma grayscale images, using NBIS matcher, are observed to be 78.82% and 78.06%, respectively. On the other hand, the GAR values for raw and inverted ordinary grayscale images are 65.52% and 61.61%, respectively.

The second matcher employed in our work, or MCC, also illustrates significant improvement in the verification performance from the Luma grayscale images. ROC curves corresponding to MCC are shown in Fig. 9 (b). The EERs of original and inverted Luma grayscale contactless fingerprints are 8.19% and 8.28%, respectively. These values are 38.18% and 41.15% higher than their counterparts from the experiments performed with ordinary grayscale images. Similarly, the GARs (@FAR=0.01%) achieved with original and inverted Luma grayscale fingerprints are 75.94% and 74.67%, respectively. These values illustrate remarkable improvements over their counterparts with the ordinary grayscale images, or 30.66% and 42.17% respectively.

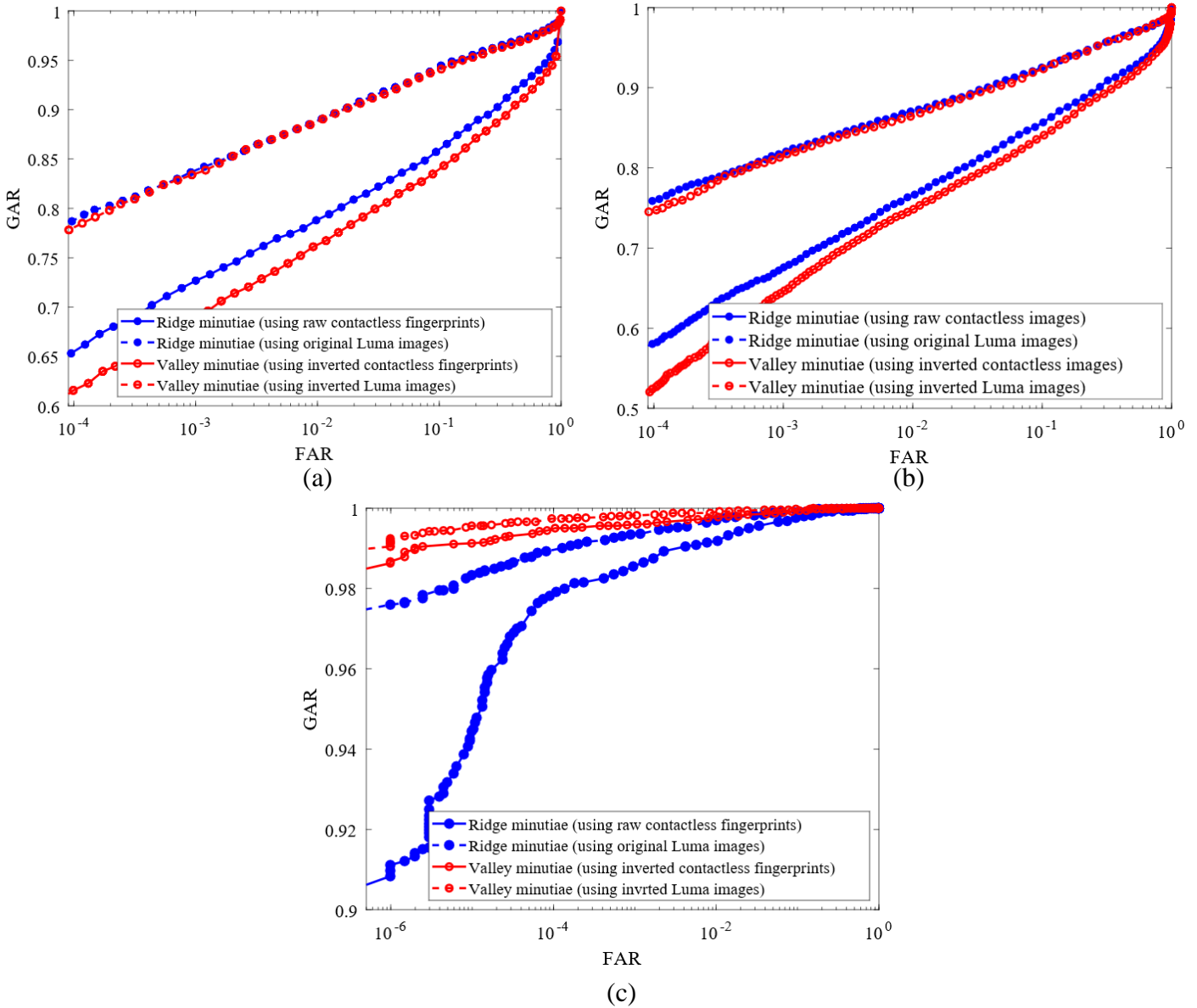

**Figure 9**: Comparative ROC curves for original and inverted contactless fingerprints using ordinary and Luma grayscale images, (a) using NBIS, (b) using MCC, (c) using COTS matcher.

Lastly, the experiments using the COTS matcher with raw Luma grayscale images achieved an EER and GAR (@FAR=0.0001%) of 0.45% and 97.6% respectively. While, same metrics for the ordinary grayscale images are 0.81% and 91.11% respectively. Therefore, it can be inferred that the Luma grayscale images are outperforming ordinary grayscale images with 44.44% and 7.12% improvements in EER and GAR, respectively. On the other hand, the performance metrics for inverted Luma grayscale images were 0.17% (EER) and 99.25% (GAR@FAR=0.0001%), respectively. While for inverted ordinary grayscale images, the corresponding EER and GAR values were 0.29% and 98.68% respectively. Therefore, the percentage improvements for inverted grayscale images are 41.37% and 0.57%, respectively for EER and GAR. The ROC curves for the COTS matcher with Luma grayscale images are shown in Fig. 9 (c). These outperforming results from Luma grayscale fingerprints clearly indicate that a more effective representation of colored images to the grayscale space can be used to achieve significantly enhanced contactless fingerprint matching.

## 5. Conclusions and Further Work

This paper presented a detailed analysis of contactless fingerprint images, on the minutiae representation and matching, to achieve significant performance improvements using popular fingerprint matchers. The experimental results presented in this paper on two public contactless fingerprint databases [13], [16], using MCC [18], NBIS [17] and COTS [12] matchers consistently illustrate significant improvement in the performance over the conventional approaches in the literature. The work detailed in this paper indicates that the grayscale polarity alterations under the ambient or indoor illumination is frequent and should be incorporated into the design of any effective matching strategy to utilize full potential from the contactless fingerprint images. Absence of contact between the sensor platen and the human finger often leads to varying illuminated areas in the fingerprint images. Therefore, the ridges in such images are often rendered as darker *and* brighter in different regions. This is unlike for the images from popular contact-based fingerprint sensors, in which the ridges are always rendered as darker and valleys as brighter, largely due to the frustrated total internal reflection. Simultaneous recovery and use of ridge and valley minutiae in the contactless fingerprints can enable superior matching capabilities by utilizing the potential from the inverted fingerprints, either individually or in combination with the raw fingerprints.

Above arguments are evaluated using three popular fingerprint matchers, namely NBIS, MCC and COTS. On one hand, NBIS and MCC yield comparable performance for both ridge (raw fingerprints) and valley (inverted fingerprints) minutiae matching approaches. However, the combination of scores from raw and inverted fingerprints achieves significantly improved performance with both of these matchers as well. Significant performance improvement, in EER and ROC or GAR, due to such combination can validate our arguments. On the other hand, COTS matcher clearly illustrates improved performance with inverted fingerprints alone, which can further be improved through the score combination. This work also considered the effective conversion of contactless color fingerprint representation to its grayscale representation which has received almost nil attention in the literature. In this context, a more diversified grayscale representation, namely Luma grayscale, was introduced with quite encouraging results. Such effective grayscale representation is quite valuable for improving the performance from contactless fingerprints acquired using the widely popular color cameras on mobile phones. Contactless fingerprint acquisition for a range of mobile application requires its detection under complex or moving backgrounds and under involuntary finger motions. Such detection can be achieved using a faster-RCNN, similar to as illustrated in [6] for palmprint acquisition and is suggested for the further work.